%% file: main.tex
\documentclass{article}

\usepackage{microtype}
\usepackage{graphicx}
\usepackage{subfigure}
\usepackage{multirow,booktabs} 
\usepackage[table,xcdraw, dvipsnames]{xcolor}
\usepackage{bbm}

\usepackage[hidelinks]{hyperref}

\usepackage[accepted]{icml2025}

\definecolor{myblue}{RGB}{0, 0, 128}
\definecolor{mygreen}{RGB}{0, 128, 0}
\definecolor{myred}{RGB}{150, 0, 0}

\definecolor{tabblue}{HTML}{1f77b4}
\definecolor{taborange}{HTML}{ff7f0e}
\definecolor{tabgreen}{HTML}{2ca02c}
\definecolor{tabred}{HTML}{d62728}
\definecolor{tabpurple}{HTML}{9467bd}
\definecolor{tabbrown}{HTML}{8c564b}
\definecolor{tabpink}{HTML}{e377c2}
\definecolor{tabgray}{HTML}{7f7f7f}
\definecolor{tabolive}{HTML}{bcbd22}
\definecolor{tabcyan}{HTML}{17becf}

\usepackage{amsmath}
\usepackage{amssymb}
\usepackage{mathtools}
\usepackage{amsthm}

\usepackage{enumitem}
\usepackage{multicol}
\usepackage{multirow}
\usepackage{wrapfig}
\usepackage{subcaption}
\usepackage{nicefrac}
\usepackage{tikz}

\newcommand{\repourl}{\url{https://github.com/andreacini/corel}}

\input{notation}

\input{acronyms}

\theoremstyle{plain}
\newtheorem{theorem}{Theorem}[section]
\newtheorem{proposition}[theorem]{Proposition}

\theoremstyle{definition}

\theoremstyle{remark}

\usepackage[textsize=tiny]{todonotes}

\makeatletter
\newcommand{\pushright}[1]{\ifmeasuring@#1\else\omit\hfill$\displaystyle#1$\fi\ignorespaces}
\newcommand{\pushleft}[1]{\ifmeasuring@#1\else\omit$\displaystyle#1$\hfill\fi\ignorespaces}
\makeatother

\icmltitlerunning{Relational Conformal  Prediction for Correlated Time Series}

\begin{document}

\twocolumn[
\icmltitle{Relational Conformal  Prediction for Correlated Time Series}

\icmlsetsymbol{equal}{*}

\begin{icmlauthorlist}
\icmlauthor{Andrea Cini}{usi,snsf}
\icmlauthor{Alexander Jenkins}{usi,icl}
\icmlauthor{Danilo Mandic}{icl}
\icmlauthor{Cesare Alippi}{usi,polimi}
\icmlauthor{Filippo Maria Bianchi}{uit,norse}
\end{icmlauthorlist}

\icmlaffiliation{usi}{IDSIA USI-SUPSI, Universit\`a della Svizzera italiana}
\icmlaffiliation{snsf}{Swiss National Science Foundation Postdoc Fellow}
\icmlaffiliation{polimi}{Politecnico di Milano}
\icmlaffiliation{icl}{Imperial College London}
\icmlaffiliation{uit}{UiT The Arctic University of Norway}
\icmlaffiliation{norse}{NORCE Norwegian Research Centre AS}

\icmlcorrespondingauthor{Andrea Cini}{andrea.cini@usi.ch}
\icmlkeywords{uncertainty quantification, time series forecasting, conformal prediction, graph deep learning, correlated time series}

\vskip 0.3in
]

\printAffiliationsAndNotice{}  

\input{abstract}
\input{content}


\input{impact_statement}
\input{acks}

\bibliographystyle{plainnat}
\bibliography{bibliography}

\clearpage

\appendix
\onecolumn
\input{appendix}

\end{document}

%% file: notation.tex
\usepackage{amsmath, amsfonts, bm, tabstackengine}

\usepackage{etoolbox}
\makeatletter
\patchcmd{\hyper@makecurrent}{%
    \ifx\Hy@param\Hy@chapterstring
        \let\Hy@param\Hy@chapapp
    \fi
}{%
    \iftoggle{inappendix}{
        \@checkappendixparam{chapter}%
        \@checkappendixparam{section}%
        \@checkappendixparam{subsection}%
        \@checkappendixparam{subsubsection}%
        \@checkappendixparam{paragraph}%
        \@checkappendixparam{subparagraph}%
    }{}%
}{}{\errmessage{failed to patch}}

\newcommand*{\@checkappendixparam}[1]{%
    \def\@checkappendixparamtmp{#1}%
    \ifx\Hy@param\@checkappendixparamtmp
        \let\Hy@param\Hy@appendixstring
    \fi
}
\makeatletter

\newtoggle{inappendix}
\togglefalse{inappendix}

\apptocmd{\appendix}{\toggletrue{inappendix}}{}{\errmessage{failed to patch}}

\newcommand{\autorefp}[1]{(\autoref{#1})}
\newcommand{\autorefseq}[2]{\autoref{#1}--\ref{#2}}

\newcommand{\gnn}{\text{GNN}}

\def\bigO{\mathcal{O}}

\newcommand{\newterm}[1]{{\emph{#1}}}

\def\eqref#1{equation~\ref{#1}}

\def\1{\bm{1}}

\def\vtheta{{\bm{\theta}}}

\def\vh{{\bm{h}}}

\def\vq{{\bm{q}}}
\def\vr{{\bm{r}}}

\def\vu{{\bm{u}}}
\def\vv{{\bm{v}}}

\def\vx{{\bm{x}}}
\def\vy{{\bm{y}}}
\def\vz{{\bm{z}}}

\def\mA{{\bm{A}}}

\def\mH{{\bm{H}}}

\def\mQ{{\bm{Q}}}
\def\mR{{\bm{R}}}

\def\mU{{\bm{U}}}
\def\mV{{\bm{V}}}

\def\mX{{\bm{X}}}
\def\mY{{\bm{Y}}}
\def\mZ{{\bm{Z}}}

\def\mPhi{{\bm{\Phi}}}

\DeclareMathAlphabet{\mathsfit}{\encodingdefault}{\sfdefault}{m}{sl}
\SetMathAlphabet{\mathsfit}{bold}{\encodingdefault}{\sfdefault}{bx}{n}

\def\gE{{\mathcal{E}}}
\def\gF{{\mathcal{F}}}

\def\gQ{{\mathcal{Q}}}
\def\gR{{\mathcal{R}}}

\def\sI{{\mathbb{I}}}

\def\sR{{\mathbb{R}}}

\DeclareMathOperator*{\argmin}{arg\,min}


%% file: acronyms.tex
\usepackage[acronym, nohypertypes={acronym}]{glossaries}

\newacronym{method}{\textsc{CoRel}}{\textit{\underline{Co}nformal \underline{Rel}ational Prediction}\nobreak}
\newacronym{gdl}{GDL}{graph deep learning}
\newacronym{cp}{CP}{\textit{conformal prediction}}
\newacronym{scp}{SCP}{{split conformal prediction}}
\newacronym{iqn}{IQN}{implicit quantile network}
\newacronym{relqn}{RelQP}{\textit{relational quantile predictor}}
\newacronym{pi}{PI}{prediction interval}

\newglossaryentry{scpi}{name=SPCI,description=}
\newglossaryentry{nexcp}{name=NexCP,description=}
\newglossaryentry{seqcp}{name=SeqCP,description=}
\newglossaryentry{hopcpt}{name=HopCPT,description=}
\newglossaryentry{cornn}{name=\textsc{CoRNN},description=}

\newacronym{stgnn}{STGNN}{spatiotemporal graph neural network\nobreak}
\newacronym{gnn}{GNN}{graph neural network}
\newacronym{mp}{MP}{message-passing}
\newacronym{mlp}{MLP}{multilayer perceptron}
\newacronym{tts}{TTS}{time-then-space}
\newacronym{stt}{STT}{space-then-time}
\newacronym{tas}{T\&S}{time-and-space}
\newacronym{rnn}{RNN}{recurrent neural network}
\newacronym{tcn}{TCN}{temporal convolutional network}
\newacronym{gru}{GRU}{gated recurrent unit}
\newacronym{sn}{SN}{\textit{sensor network}}
\newacronym{iot}{IoT}{Internet of Things}
\newacronym{stmp}{STMP}{spatiotemporal message-passing}
\newacronym{stcn}{STCN}{spatiotemporal convolutional network}
\newacronym{gcrnn}{GCRNN}{graph convolutional recurrent neural network}
\newacronym{mae}{MAE}{mean absolute error}
\newacronym{mape}{MAPE}{mean absolute percentage error}
\newacronym{rmse}{RMSE}{root mean squared error}
\newacronym{mre}{MRE}{mean relative error}

\glsdisablehyper
\newglossaryentry{ttsimp}{name=RNN\texttt{+}IMP,description=}
\newglossaryentry{ttsamp}{name=RNN\texttt{+}AMP,description=}
\newglossaryentry{tasimp}{name=GCRNN-IMP,description=}
\newglossaryentry{tasamp}{name=GCRNN-AMP,description=}

\newglossaryentry{fcrnn}{name=FC-RNN,description=}
\newglossaryentry{local_rnn}{name=LocalRNNs,description=}
\newglossaryentry{dcrnn}{name=DCRNN,description=}
\newglossaryentry{agcrn}{name=AGCRN,description=}
\newglossaryentry{gwnet}{name=GraphWaveNet,description=}

\newglossaryentry{gpvar}{name=GPVAR,description=}
\newglossaryentry{lgpvar}{name=GPVAR-L,description=}
\newglossaryentry{la}{name=METR-LA,description=}
\newglossaryentry{bay}{name=PEMS-BAY,description=}
\newglossaryentry{cer}{name=CER-E,description=}
\newglossaryentry{air}{name=AQI,description=}
\newglossaryentry{pems3}{name=PEMS03,description=}
\newglossaryentry{pems4}{name=PEMS04,description=}
\newglossaryentry{pems7}{name=PEMS07,description=}
\newglossaryentry{pems8}{name=PEMS08,description=}

%% file: abstract.tex
\begin{abstract}
We address the problem of uncertainty quantification in time series forecasting by exploiting observations at correlated sequences. 
Relational deep learning methods leveraging graph representations are among the most effective tools for obtaining point estimates from spatiotemporal data and correlated time series.
However, the problem of exploiting relational structures to estimate the uncertainty of such predictions has been largely overlooked in the same context. To this end, we propose a novel distribution-free approach based on the conformal prediction framework and quantile regression. Despite the recent applications of conformal prediction to sequential data, existing methods operate independently on each target time series and do not account for relationships among them when constructing the prediction interval. We fill this void by introducing a novel conformal prediction method based on graph deep learning operators. 
Our approach, named \acrfull{method}, does not require the relational structure (graph) to be known a priori and can be applied on top of any pre-trained predictor. Additionally, \acrshort{method} includes an adaptive component to handle non-exchangeable data and changes in the input time series. Our approach provides accurate coverage and achieves state-of-the-art uncertainty quantification in relevant benchmarks.
\end{abstract}

%% file: content.tex
\section{Introduction}\label{sec:intro}

Many recent advancements in deep learning methods for time series forecasting rely on learning from large collections of~(related) time series~\cite{benidis2022deep, liang2024foundation}. In many application domains, such time series are characterized by a rich spatiotemporal dependency structure that can be exploited by introducing inductive biases in the forecasting architecture~\cite{cini2023graphdeep}, to steer the learning procedure toward the most plausible models. Accounting for the existing dependencies, usually represented as a graph, allows the resulting models to obtain accurate predictions with a reduced sample complexity~\cite{jin2023survey, cini2023graphdeep}. Besides the accuracy of the point estimates, the \textit{reliability} of the forecasts is a critical aspect of the problem and a key element to enable effective decision-making in many applications~\cite{makridakis1996forecasting, petropoulos2022forecasting}.
Uncertainty quantification methods~\cite{smith2024uncertainty, vovk2005algorithmic} can improve reliability by providing confidence intervals on the forecasting error magnitude, allowing for making more informed decisions~\cite{hyndman2018forecasting}. This is particularly true for risk-sensitive applications such as healthcare~\cite{makridakis2019forecasting} and load forecasting~\cite{gasparin2022deep}. In this context, inter-series~(spatiotemporal) dynamics offer both a challenge and an opportunity. Indeed, while these dependencies can lead to wide \glspl{pi} if overlooked, they may also provide additional knowledge to reduce uncertainty~\cite{zambon2022aztest}.

Existing probabilistic forecasting frameworks often rely on strong distributional assumptions and major modifications of the base point predictor~\cite{benidis2022deep, salinas2020deepar}. As such, they cannot be used to quantify uncertainty given a pre-trained forecasting model. 
In such a setting, \gls{cp}~\cite{vovk2005algorithmic, angelopoulos2023conformal} methods are particularly appealing. \Gls{cp} is an uncertainty quantification framework that estimates confidence intervals with marginal coverage guarantees from observed prediction residuals. 
One of the main assumptions of standard \gls{cp} methods is that of exchangeability between the data used to estimate the confidence intervals and the test data points, i.e., the assumption that their joint probability distribution is invariant to the ordering of the associated sequence of random variables~\cite{angelopoulos2024theoretical}. Although this assumption does not usually hold when operating on time series~\cite{barber2023conformal}, several methods have successfully adapted \gls{cp} to estimate forecast uncertainty~\cite{stankeviciute2021conformal, xu2023conformal, xu2023sequential, jensen2022ensemble, auer2023conformal}. Nevertheless, existing \gls{cp} approaches operate on each (possibly multivariate) time series independently and cannot account for dependencies among correlated time series. 

In this paper, we propose \gls{method}, a novel \gls{cp} approach leveraging graph representations and \gls{gdl} for quantifying uncertainty in correlated time series forecasting. In our framework, a \gls{stgnn}~\citep{jin2023survey, cini2023graphdeep} is trained on a calibration set to approximate the quantile function of the distribution of prediction residuals. Relationships among time series, assumed to be \textit{sparse}, are learned end-to-end from the observed residuals owing to a graph structure learning module integrated into the processing. Our approach estimates the error quantile function for each time series at each time step, by conditioning the shared uncertainty quantification model on past observations at neighboring nodes~(as defined by the learned graph structure). Finally, an adaptive component is added to handle potential non-stationarities by relying on a small set of parameters specific to each time series. Our approach can be applied to the residuals generated by \textit{any} point forecasting model, even those that completely disregard potential relationships among the input time series.

Our main novel contributions can be summarized as follows.
\begin{itemize}
    \item The first application of \gls{gdl} to \gls{cp} for time series;
    \item A novel, sound, and effective \gls{cp} method able to quantify uncertainty from observations across a collection of correlated time series;
    \item A family of graph-based architectures to estimate uncertainty that shares most of the learnable parameters among the processed time series, while including node-level parameters that dynamically adapt to changes in each target sequence.
\end{itemize}
Empirical results show that \gls{method} achieves state-of-the-art performance compared to existing \gls{cp} approaches for time series in several datasets and under different scenarios.

\section{Preliminaries}\label{sec:preliminaries}

This section introduces the problem settings and the preliminary concepts that serve as foundations for our approach.

\subsection{Problem Formulation}\label{sec:forecasting} 

Consider a collection of $N$ correlated time series. Denote by $\vx^i_t \in \sR$ the scalar target variable associated with the $i$-th time series at time step $t$; $\mX_t \in \sR^{N\times 1}$ indicates the $N$ stacked target variables w.r.t.\ the entire time series collection. $\mX_{t:t+T}$ indicates the sequence within time interval $[t, t+T)$; conversely, with the shorthand $\mX_{<t}$ refers to observations up to time step $t$~(excluded). Time series are assumed to be \newterm{homogenous}, i.e., all the variables~(observables) describe the same physical quantity~(e.g., temperature or energy consumption). Analogously, $\mU_t \in \sR^{N \times d_u}$ indicates the $d_u$-dimensional exogenous covariates associated with each time series. 
We assume that the $i$-th time series is generated by a stochastic \textit{time-invariant} process such as
\begin{equation}
    \vx_t^i \sim p\left(\vx_{t}^i | \mX_{<t}, \mU_{< t}\right).
\end{equation}
Let us hypothesize the existence of a \textit{sparse} predictive causality \textit{\`a la Granger}~\citep{granger1969investigating}, i.e., we assume that the values of a single time series are related to the values of a~(\textit{small}) subset of other time series in the collection. The extension of the framework to collections of multivariate time series is orthogonal to the proposed approach~(e.g., see \citealt{feldman2023calibrated}); we focus on the univariate case to maintain a contained scope. The problem of dealing with non-stationary processes will be discussed in \autoref{sec:adaptive-inference}.

\paragraph{Forecasting} We are interested in a model that produces point forecasts by predicting the unknown $H$-steps-ahead~($H\geq 0$) observation $\mX_{t+H}$ given a window $W \geq 1$ of past observations $\mX_{t-W:t}$ and the associated exogenous variables $\mU_{t-W:t}$ as
\begin{equation}
    \widehat \mX_{t+H} = \gF_\vtheta(\mX_{t-W:t}, \mU_{t-W:t}).
\end{equation}
$\gF_\theta$ denotes a generic parametric model family, i.e., a simple \gls{rnn} for univariate time series.
Given a trained model, our objective is to build a confidence interval around predictions $\widehat \mX_{t+H}$. Note that the following easily extends to multi-step predictions $\widehat \mX_{t:t+H}$, but we focus on forecasting the single time step $H$ to simplify the presentation and discussion.

\paragraph{Uncertainty quantification} Our objective is to estimate \glspl{pi}, $ C^\alpha_{i,t}(\widehat \mX_{t+H})$, such that
\begin{equation}\label{eq:marginal_coverage}
    P\left(\vx^i_{t+H} \in C^\alpha_{i,t}\left(\widehat \mX_{t+H} \right)\right) \geq 1- \alpha,
\end{equation}
where $\alpha$ is the desired confidence level. If the interval satisfies \autoref{eq:marginal_coverage}, we say that the \gls{pi} achieves marginal coverage $1-\alpha$. Similarly, we say that the \gls{pi} provides conditional coverage $1-\alpha$ if
\begin{equation}\label{eq:conditional_coverage}
    P\left(\vx^i_{t+H} \in C^\alpha_{i,t}\left(\widehat \mX_{t+H} \right) \Big| \mX_{<t}, \mU_{<t} \right) \geq 1- \alpha.
\end{equation}
Conditional coverage provides stronger guarantees and it is often harder to achieve~\cite{angelopoulos2024theoretical}. In the following, we will omit the dependence of the interval on the forecasts and simply write $C^\alpha_{i,t}$. Among uncertainty quantification methods, we are interested in post-hoc approaches that can build confidence intervals for any given pre-trained point predictor $\gF_\theta$ without requiring any modification of the base forecasting architecture.

\subsection{Conformal Prediction}\label{sec:cp}

As anticipated in \autoref{sec:intro}, standard \Gls{cp} methods~\cite{vovk2005algorithmic, angelopoulos2023conformal} are a class of distribution-free uncertainty quantification techniques that build \glspl{pi} from empirical quantiles of \textit{conformal scores}. In the forecasting setting, we consider as conformal scores the prediction residuals,
\begin{equation}\label{eq:residual}
    \vr^i_t = \vx^i_t - \hat \vx^i_t,
\end{equation}
and use $\mR_t$ to denote residuals w.r.t.\ the entire time series collection. Under appropriate assumptions, \gls{cp} methods can build valid and informative \glspl{pi}~\cite{angelopoulos2023conformal, barber2023conformal}. \Gls{scp}~\cite{vovk2005algorithmic} is arguably the most common approach and exploits scores computed on a \textit{calibration set} that is disjoint from the training data (i.e., a post-hoc approach). 

As mentioned, most standard \gls{cp} methods rely on the assumption that calibration and test data are \textit{exchangeable}, which allows the procedure to treat them symmetrically and obtain valid~(marginal) coverage guarantees~\cite{angelopoulos2024theoretical}. Since this assumption does not hold when dealing with time series data, there have been several recent results extending the \gls{cp} framework beyond exchangeability~\cite{tibshirani2019conformal,stankeviciute2021conformal, gibbs2021adaptive, xu2023conformal}. In particular, \citet{barber2023conformal} showed that approximate coverage can be achieved by reweighting the residuals to account for the lack of exchangeability between calibration and test set. \citet{auer2023conformal} learn such a reweighting scheme through an attention-based architecture. Differently, \citet{xu2023sequential} introduce \gls{scpi}, a method based on fitting a quantile random forest~\cite{meinshausen2006quantile} on the most recent prediction residuals at each time step. Similar to \gls{scpi}, our approach relies on quantile regression to build \glspl{pi} but differently from existing methods, it exploits observations in arbitrary sets of time series by relying on \gls{gdl} operators. 

\subsection{Quantile Regression}\label{sec:quantile-reg}

Quantile regression~\cite{koenker2001quantile} is an established statistical framework that consists of learning a model of the quantile function~(the inverse c.d.f.) of a target distribution from observations. In particular, given $\vy \sim p(\vy | \vx)$ and observations $(x_1, y_1), \dots, (x_N, y_N)$, a standard approach to estimate the $\alpha$-quantile is to train a model by minimizing the so-called pinball loss
\begin{equation}\label{eq:pinball}
    \ell^\alpha(\hat q^\alpha(x), y) = 
    \begin{cases}
        (1-\alpha)(\hat q^\alpha(x) - y), \, &\hat q^\alpha(x) \geq y\\
        \alpha(y - \hat q^\alpha(x)), \, &\hat q^\alpha(x) < y
    \end{cases}
\end{equation}
where $\hat q^\alpha(x)$ is the estimate of the $\alpha$-quantile w.r.t.\ $x$. 

\paragraph{Quantile networks} Quantile regression has been incorporated in several probabilistic forecasting architectures~\cite{benidis2022deep}. The simplest approach consists of using a multi-output network to predict a set of quantiles of interest and interpolate among them to approximate the entire quantile function~\cite{wen2017multi}. More complex approaches rely on, e.g.,  splines~\cite{gasthaus2019probabilistic}. Conversely, \glspl{iqn}~\cite{dabney2018implicit, ostrovski2018autoregressive, gouttes2021probabilistic} approximate the quantile function by being trained to minimize the loss in \autoref{eq:pinball} given the quantile level $\alpha$ as input and sampling a random $\alpha$ for each sample in a mini-batch. 

\input{imgs/figure_1}

\subsection{Graph Deep Learning for Time Series Forecasting}\label{sec:stgnn}

\Glspl{gnn}~\cite{bacciu2020gentle, bronstein2021geometric} process graph-structured data by incorporating the graph topology as an inductive bias, e.g., by relying on message-passing layers~\cite{gilmer2017neural}. \Glspl{stgnn}~\cite{jin2023survey, cini2023graphdeep} leverage message-passing layers within sequence modeling architectures to process spatiotemporal data and collections of time series where dependencies are represented as a (possibly dynamic) graph. We consider as reference architectures \gls{tts} models~\cite{ gao2021equivalence, cini2023graphdeep} where each time series in the collection is processed independently from the others by a temporal encoder whose output is then fed into a stack of \gls{gnn} layers. In particular, we adopt the following template architecture:
\begin{align}
    \vh^{i,0}_t &= \textsc{SeqEnc}\left(\vx^i_{t-W:t}, \vu^i_{t-W:t}\right),\\
    \mH_t^{l+1} &= \gnn_l(\mH_t^l,\mA),\label{eq:mp}, \quad l=0,\dots,L-1 \\
    \hat \vy^i_{t} &= \textsc{Readout}\left(\vh_t^{i,L}\right),
\end{align}
where $\mA \in \sR^{N\times N}$ is the graph adjacency matrix and $\hat \vy^i_{t}$ a generic node-level prediction associated with the problem at hand. $\textsc{SeqEnc}({}\cdot{})$ and $\gnn_l({}\cdot{})$ denote, respectively, any sequence modeling architecture, e.g., an \gls{rnn}, and any \gls{gnn} layer, e.g., based on message-passing. Representations can then be mapped into predictions $\widehat \mY_{t}$ by using any readout block, e.g., a \gls{mlp}. \Glspl{stgnn} have been used as forecasting architecture~($\mY_t = \mX_{t+H}$) with great success. In the following, we will exploit this framework as a backbone for estimating the residual quantile distribution. We refer to \autoref{sec:related} and \citet{jin2023survey} for more discussion on the application of \glspl{stgnn} in the context of time series analysis.

\section{Conformal Relational Prediction}\label{sec:method}

Our objective is to build \glspl{pi} by exploiting relational dependencies across the residuals of the target time series. We model the dependencies as edges of a graph and learn them under the assumption that the relational structure is \textit{sparse}, which reduces the computational costs and act as an inductive bias on the structure learning architecture. By relying on such representation, we can leverage \gls{gdl} methods for time series to process the data. In particular, we train a \gls{stgnn} on the residuals of the calibration set to predict the quantiles of the error distribution. Conditioning the prediction on the recent history of related time series allows for taking the dependency structure of the data into account when estimating uncertainty: a key aspect in applying conformal prediction to non-exchangeable data~\citep{barber2023conformal}. Compared to existing methods~\cite{xu2023sequential, auer2023conformal} that only capture temporal dependencies, our approach allows for modeling spatiotemporal dependencies among different time series. 
\autoref{sec:conformal-procedure} presents the details of the proposed conformal inference procedure by assuming that the relational structure at each time step is defined by an adjacency matrix $\mA \in \sR^{N\times N}$~\autorefp{sec:conformal-procedure}. We then show how to learn the graph structure directly from data and make the model adaptive in \autoref{sec:structure-learning} and \autoref{sec:adaptive-inference}, respectively. Finally, we discuss the theoretical properties of the approach in \autoref{sec:discussion}. \autoref{fig:corel} shows an overview of the architecture.

\subsection{Relational Quantile Predictor}\label{sec:conformal-procedure}

Consider a standard \gls{scp} setup, where the training data are split into training and calibration sets. For the moment, we disregard possible nonstationarities in the data considering the problem setup introduced in \autoref{sec:forecasting} and encode spatial dependencies in the adjacency matrix $\mA \in \sR^{N\times N}$. While the training set is used to fit the point predictor $\gF_\theta$, we use the prediction residuals in the calibration set~($\gR^{cal}$) to learn the quantile function of the error distribution at each step. 

\paragraph{Relational quantile regression} We implement the quantile regressor as a hybrid global-local \gls{stgnn}, which mixes global (shared) parameters with local, target-specific components~\cite{smyl2020hybrid}. Sharing most learnable parameters across all time series reduces sample complexity, while local parameters allow for tailoring the processing to each series. Specifically, we keep all processing blocks shared and associate a learnable node embedding $\vv^i \in \sR^{d_v}$ with each time series~\cite{cini2023taming}.
More specifically, our model is a quantile network (see \autoref{sec:quantile-reg}) composed of the following processing layers:
\begingroup
\allowdisplaybreaks
\begin{align}
    \vh^{i,0}_{t} &= \textsc{Enc}\left(\vr^i_{t-1},\vv^i\right),\label{eq:enc}\\
    \mZ_{t} &= \textsc{STGNN}\Big(\mH^{0}_{
    \leq t}, \mA\Big),\label{eq:stmp-block}\\
    \hat \vq^{i,\alpha}_{t+H} &= \textsc{QDec}\left(\alpha, \vz_{t}^{i}, \vv^i\right)\label{eq:dec},
\end{align}
\endgroup
where $\vr^i_{t-1}$ are prediction residuals~(\autoref{eq:residual}) and  $\hat \vq^{\alpha,i}_{t+H}$ is the predicted $\alpha$-quantile at time step $t+H$ for the $i$-th time series. $\textsc{Enc}({}\cdot{})$ denotes any encoding layer, e.g., a linear transformation or an \gls{mlp}. 
For the \gls{stgnn} block, several designs are possible~(e.g., see \citealt{jin2023survey}); the one we follow is the template in \autoref{sec:stgnn}.
$\textsc{QDec}({}\cdot{})$ is a readout mapping the representations at each node to the prediction of the quantile of specified level $\alpha$. 
We refer to the family of quantile networks defined in \autorefseq{eq:enc}{eq:dec} as \glspl{relqn} and use the notation
\begin{equation}\label{eq:quantile-estimator}
    \widehat \mQ^\alpha_t = \gQ_{\psi}\left(\alpha, \mV; \mR_{t-W:t}, \mA\right),
\end{equation}
where $\gQ_{\psi}$ indicates the shared~(global) part of the network and $\widehat \mQ^\alpha_t \in \sR^N$ denotes the predicted $\alpha$-quantiles at time step $t$ w.r.t.\ the full time series collection. Note that the framework can easily accommodate further inputs at the encoding block~(e.g., we can condition the regression on $\mX_{<t}$ and $\mU_{<t}$). 
The model is trained by minimizing the pinball loss~(\autoref{eq:pinball}) at each time step in the calibration set w.r.t.\ the full-time series collection. 
Through the message-passing layers, the residuals of each time series contribute to estimating the quantiles of the error distribution at neighboring nodes. In practice, we restrict the input of the regressor to the most recent observations rather than considering the full sequence~(the window length here can also be different from the one used by the point predictor). 

\paragraph{Building the confidence intervals} Given the trained quantile network $\gQ_\psi$, we build the \glspl{pi} for each target~(test) time step as
\begin{equation}\label{eq:cp-interval}
    \widehat C^\alpha_{i,t} = \left[\hat \vx^i_{t+H} + \hat{\vq}^{i,\alpha/2}_t, \hat \vx^i_{t+H} + \hat{\vq}^{i,1-\alpha/2}_t\right],
\end{equation}
or
\begin{align}
    \hat \beta_i &= \argmin_{\beta_i} \Big|\hat{\vq}^{i,1-\alpha/2 + \beta_i}_t - \hat{\vq}^{i,\alpha/2 + \beta}_t\big|\\
    \widehat C^\alpha_{i,t} &= \left[\hat \vx^i_{t+H} + \hat{\vq}^{i,\alpha/2 + \hat \beta_i}_t, \hat \vx^i_{t+H} + \widehat{\vq}^{i,1-\alpha/2 + \hat \beta_i}_t\right],\label{eq:cp-interval-optimized}
\end{align}
where  $\widehat C^\alpha_{i,t}$ indicates the estimated \gls{pi}. While both \autoref{eq:cp-interval} and \autoref{eq:cp-interval-optimized} correspond to the same confidence level, the \gls{pi} in \autoref{eq:cp-interval-optimized} can be narrower, at the expense of the additional computation needed to obtain $\hat \beta$~\cite{xu2023conformal}. In practice, one can choose between the two approaches given computational constraints and the difference in performance observed on a validation set.  Note that $\widehat C^\alpha_{i,t}$ can provide only \textit{approximate} coverage, as it is subject to approximation errors of the true quantile function. \autoref{sec:discussion} will discuss this aspect in detail. 
A potential drawback of \gls{method} is that residuals cannot be assumed exchangeable in most practical scenarios. The error distribution can be \textit{non-stationary}, making it difficult to obtain any coverage guarantee. To mitigate the problem, \autoref{sec:adaptive-inference} discusses an efficient and scalable approach to make the framework adaptive by updating local components of the architecture over time. Finally, it is worth noting that the use of relational components in \gls{method} relies on the actual presence of the associated dependencies in the data. 
In practical applications, the presence of spatiotemporal correlations in the residuals can be verified through ad-hoc statistical tests ~\cite{zambon2022aztest, zambon2023where}, whose outcome can support the adoption of \gls{method}. 

\subsection{Learning the Relational Structure}\label{sec:structure-learning}

Assuming the dependency structure across time series to be unknown, we integrate a graph learning module into the architecture to derive the operational graph topology directly from the residuals. To do so, we adopt a probabilistic structure learning framework~\cite{niculae2023discrete, cini2023sparse, manenti2024learning}. In particular, we associate each edge with a score $\phi^{ij}$ and learn a distribution over $K$-NN graphs parametrized by the matrix $\mPhi \in \sR^{N\times N}$~\cite{cini2023sparse, kazi2022differentiable}. Notably, we consider graphs obtained by sampling, for each $i$-th node, $K$ elements \textit{without replacement} from the categorical distribution 
\begin{align}\label{eq:graph-sampler}
    \mPhi &= \gE_{\bm{\xi}}\left(\mR_{<t}, \mV, \dots\right)\\
    M_i &= \text{Categorical}\left(\frac{\exp\{\phi^{ik}\}}{\sum_{j=1}^N\exp\{\phi^{ij}\}};k \in \{1, \dots, N\}\right),
\end{align}
where $\gE_{\bm{\xi}}({}\cdot{})$ is a generic trainable encoder with parameters $\bm{\xi}$. In practice, sampling can be done efficiently by exploiting the $\text{GumbelTopK}$ trick~\cite{kool2019stochastic} and scores $\mPhi$ can be parametrized directly as $\mPhi = \bm{\xi}$. 

\paragraph{End-to-end learning} To propagate gradients through the sampling, we rely on the continuous relations introduced by ~\citet{xie2019reparameterizable} paired with a straight-through gradient estimator~\cite{bengio2013estimating} to obtain discrete samples. Optionally, we sparsify the gradients by backpropagating only through a random subset of the zero entries of $\mA$~({more details are provided in \autoref{a:corel}}). As already mentioned, different parametrizations and gradient estimators for subset samplers exist and can be considered. Furthermore, if the sparsity assumption is deemed unrealistic for the problem at hand, other distributions~(e.g., based on Bernoulli random variables) can be considered~\cite{cini2023sparse}. 

\subsection{Theoretical Analysis and Further Discussion}\label{sec:discussion}

We start the discussion by providing an intuitive bound on the approximate coverage provided by \gls{method}. 

\begin{proposition}\label{p:coverage-bound} Let $P^{c}_{t+H}(\vx^i_{t+H}) = p_{t+H}(\vx^i_{t+H} \mid \mX_{<t}, \mU_{<t})$ and $P^c_{\psi}(\vx^i_{t+H}) = p_{\psi}(\vx_{t+H} \mid \mX_{<t},\mU_{<t})$ be the true conditional data-generating distribution at the test point $t+H$ and the probability distribution associated with the learned quantile function $\gQ_{\psi}$, respectively. Then
\begin{align*}
    P^c_{t+H}\left(\vx^i_{t+H} \in \widehat C^{\alpha}_{i,t} (\widehat \mX_{t+h})\right) \geq 1 - \alpha - {TV}\left(P^c_{\psi}, P^c_{t+H}\right)
\end{align*}
where ${TV}({}\cdot{})$ denotes the total variation function.
\end{proposition}
The proof relies on the properties of the total variation of probability measures and can be found in \autoref{a:theory}. Here, differently from the problem settings introduced in \autoref{sec:forecasting}, we do not assume the process to be time-invariant. \autoref{p:coverage-bound} links the \textit{conditional} coverage gap to the approximation error in estimating the quantile function of the residuals. The bound provided in \autoref{p:coverage-bound} shares similarities with the one in \cite{barber2023conformal}, which bounds the miscoverage gap for \gls{cp} from weighted empirical quantiles. \autoref{p:coverage-bound} can be seen as an analogous result that holds when estimates obtained from empirical quantiles are replaced with a parametric function approximation. By making assumptions on the expressivity of the quantile regressor in \autoref{eq:quantile-estimator} and on the stationarity process~(e.g., by assuming a strongly mixing process), we can expect the total variation between the learned and true distribution to shrink asymptotically as the size of the calibration set increases. Moreover, in this case, monitoring the coverage gap on a validation set offers an estimate of the actual miscoverage on test data. Similar analyses have been carried out for recently introduced \gls{cp} methods for time series \cite{xu2023sequential,lee2025kernelbased}; we refer the reader to these related works. If we instead expect the process to be non-stationary, $\gQ_\psi$ has to be updated over time to keep the coverage gap contained. Within this context, the next section discusses a simple and sample-efficient approach to make \gls{method} adaptive. Finally, the computational complexity of \gls{method} will depend on the \gls{stgnn} used to implement the \gls{relqn} and the number of edges sampled while learning the relational structure. For example, the cost of a forward pass for a standard \gls{tts} \gls{stgnn} with window size $W$ and a graph with $E$ edges would scale as $\bigO(WN + E)$~\cite{cini2023graphdeep}.

\subsection{Adaptation}\label{sec:adaptive-inference}

The \gls{relqn} model introduced in \autoref{sec:conformal-procedure} can yield arbitrarily large coverage gaps in the presence of distribution shifts from the calibration set where the model is trained.
Adopting a re-training approach, such as in \citet{xu2023sequential}, would be impractical due to the higher sample complexity entailed by the deep learning approach that we adopt. 
Therefore, to mitigate this issue while keeping the computational complexity under control, we update only the local components of the model over time, i.e., the learnable node embeddings $\mV$~\cite{cini2023taming}. This allows for keeping most of the learnable parameters fixed and fine-tuning only a small number of weights for each node. Empirically, we show that this procedure can effectively improve the quality of the uncertainty estimates.

\section{Related Work}\label{sec:related}

The problem of quantifying forecast uncertainty is central in fundamental and applied research in time series forecasting~\cite{hyndman2018forecasting, petropoulos2022forecasting}. Among deep learning approaches~\cite{benidis2022deep}, many generative architectures have been proposed as means to obtain probabilistic forecasts~\cite{salinas2020deepar,rangapuram2018deep, debezenac2020normalizing, rasul2021autoregressive}. Most related to our approach are those methods that exploit quantile regression~\cite{wen2017multi, gasthaus2019probabilistic, kan2022multivariate, gouttes2021probabilistic}. 
Similarly to \gls{method}, these quantile regression techniques do not usually require strong assumptions on the data distribution.  

\paragraph{Uncertainty quantification in \acrshortpl{stgnn}}  Regarding probabilistic graph-based forecasting architecture, the existing literature is limited~\cite{jin2023survey, cini2023graphdeep}. \citet{wu2021quantifying} investigate the combination of \glspl{stgnn} with standard uncertainty quantification techniques for deep learning. \citet{pal2021rnn} use an \gls{stgnn} to implement a state-space model and quantify uncertainty within a Bayesian framework. \citet{wen2023diffstg} propose a probabilistic predictor based on combining \glspl{stgnn} with a diffusion model~\cite{ho2020denoising}. \citet{zambon2023graph} introduce a framework for designing probabilistic graph state-space models that can process collections of time series. However, all these methods cannot operate on top of an existing pre-trained model and require training an ad-hoc forecasting model. Conversely, \gls{method} is trained, within a \gls{cp} framework, on predicting the quantiles of the error distribution of any existing model, rather than on forecasting the target variable.

\paragraph{Conformal prediction} Related work on \gls{cp} for time series  has been already discussed in \autoref{sec:cp} and \autoref{sec:method}. Related to our method, \citet{mao2024valid} propose a \gls{cp} approach for~(static) spatially correlated data. \citet{jiang2024spatio} propose to quantify the uncertainty in predicting power outages by fitting a quantile random forest~\cite{meinshausen2006quantile} on time series from neighboring geographical units.
\gls{method} can be framed among the \gls{cp} methods that learn a model of conformal scores distribution~\citep{xu2023sequential, lee2024conformal}. Unlike existing methods that operate on each time series separately, the estimates are conditioned on errors at both the target time series as well as at neighboring nodes.
To the best of our knowledge, no previous \gls{cp} method has been designed to specifically operate on collections of correlated time series and exploit graph deep learning operators. \gls{cp} methods for multivariate time series do exist~\cite{xu2024conformal, sun2024copula, feldman2023calibrated}, but operate on a single multidimensional time series. Moreover, although global-local models are popular among forecasting architectures~\cite{smyl2020hybrid, benidis2022deep}, \gls{method} is the first \gls{cp} architecture of this kind. Finally, \gls{cp} methods have also been applied to \textit{static} graphs and used to quantify the uncertainty of \glspl{gnn}, both in inductive~\cite{zargarbashi2023conformal} and transductive~\cite{huang2024uncertainty} settings. These methods often assume node/edge exchangeability~\cite{zargarbashi2023conformal, huang2024uncertainty} or are limited to node classification~\cite{clarkson2023distribution} or link prediction~\cite{zhao2024conformalized}. Recently, \citet{davis2024valid} proposed a \gls{cp} method for node classification with \glspl{gnn} in dynamic networks.

\input{tables/table-benchmarks}

\section{Experiments}\label{sec:experiments}

We validate \gls{method} across three experimental settings. In the first one~(\autoref{sec:benchmarks}), we compare it against state-of-the-art \gls{cp} methods operating on the residuals produced by different forecasting models. Then, we analyze \gls{method} in a controlled environment~(synthetic dataset). Finally, we assess the effectiveness of the procedure described in \autoref{sec:adaptive-inference} in adaptively improving the \glspl{pi}.
We implement \textbf{\gls{method}} as an \gls{rnn} followed by two message-passing layers. To approximate the quantile function, we train the model by minimizing the pinball loss over a discrete set of quantiles, similarly to \citet{wen2017multi}. \glspl{pi} are constructed as in \autoref{eq:cp-interval}; \autoref{a:beta} shows results for the alternative construction in \autoref{eq:cp-interval-optimized}. To learn the graph, we directly parametrize the score matrix $\mPhi$ by associating a learnable parameter with each of its entries.
We use as metrics: 1)~the difference between the specified confidence level $1-\alpha$ and the observed coverage on the test set~(\textit{$\Delta$Cov}), 2)~the width of the \gls{pi}~(\textit{\gls{pi}-Width}), and 3)~the Winkler score~\cite{winkler1972decision}, which is computed as the width of the \gls{pi} plus penalty for each observation outside of the predicted interval proportional to the actual error~(\textit{Winkler}). Note that balancing coverage and \gls{pi} width is the main challenge. More details and results are provided in the appendix.

\subsection{Time Series Forecasting Benchmarks}\label{sec:benchmarks}\label{s:benchmark}

We consider the following datasets, each coming from a different application domain: 
\textbf{\gls{la}} from the traffic forecasting literature~\cite{li2018diffusion}; a collection of air quality measurements from different Chinese cities~(\textbf{\gls{air}})~\cite{zheng2015forecasting}; a collection of energy consumption profiles acquired from smart meters within the CER smart metering project~(\textbf{\gls{cer}})~\cite{cer2016cer, cini2022filling}. 
We follow the preprocessing steps of previous works~\cite{li2018diffusion, wu2019graph, cini2023taming} and adopt $40\%/40\%/20\%$ splits for training, calibration, and testing, respectively. 
For each dataset, we first train $3$ different baseline models: a simple \textbf{\gls{rnn}} with \gls{gru} cells~\cite{cho2014properties}, a decoder-only \textbf{Transformer}~\cite{vaswani2017attention}, and a simple \gls{tts} \textbf{\gls{stgnn}} obtained by following the template in~\autoref{sec:stgnn}.
The latter uses a pre-defined graph that models the dependencies across the time series.
After training, we evaluate each baseline on the calibration set and save the associated residuals, which are then used as input to the different \gls{cp} methods. More details on the datasets and base models are provided in \autoref{a:datasets}.

\paragraph{Baselines} We compared \gls{method} against the following baselines: 1) \textbf{\gls{scp}}, the standard split \gls{cp}; 2) \textbf{\gls{seqcp}}, where, analogously to \citet{xu2023conformal}, 
we compute empirical quantiles using only the most recent $K$ residuals at each time step; 3) \textbf{\gls{nexcp}}~\cite{barber2023conformal}, which computes empirical quantiles by assigning exponentially decaying weights to past residuals; 4) \textbf{\gls{scpi}}~\cite{xu2023sequential}, which estimates the residuals' quantile function from the last few steps of each time series with a random forest; 5) \textbf{\gls{hopcpt}} which reweights past residuals by learning attention scores with a Modern Hopfield
Network~\cite{ramsauer2021hopfield}. 
These baselines are representative of the current state-of-the-art in \gls{cp} for time series forecasting; in particular, \gls{scpi}~(based on quantile regression) and \gls{hopcpt}~(based on reweighting) are representative of the main recent paradigms. Note that a comparison with non-post-hoc methods would be problematic, as results heavily depend on the base predictors being used. 
We also include in our comparison a model called \textbf{\gls{cornn}}, where we use the same architecture as \gls{method} but remove message-passing layers and node embeddings. \Gls{cornn} is an ablation of the introduced designs. 
Except for \gls{hopcpt}, which uses a custom procedure~\cite{auer2023conformal}, model selection is performed on a validation set by optimizing the Winkler score.

\paragraph{Results} \autoref{tab:exp1} reports the results across the datasets and the base prediction models. The first observation is that \gls{method} outperforms the competitors in terms of Winkler score in almost all cases. 
We observed a few exceptions only when the baseline is itself an \gls{stgnn}, as it is already expected to take care of modeling spatiotemporal dependencies. 
However, note that the \gls{stgnn} base model has access to a pre-defined graph, which is not always available in practical applications. 
In terms of coverage, \gls{method} achieves good results, with the exception of some cases in the \gls{cer} dataset. However, we note that our model selection prioritized the Winkler score which emphasizes the width of the prediction bands besides the coverage. \gls{cornn}, the simplified version of our approach, obtains good overall performance and is competitive against the state-of-the-art but, despite providing good coverage, it is outperformed by \gls{method} in most scenarios in terms of Winkler score. 
Among the competitors, \gls{hopcpt} provides competitive results in \gls{la} and \gls{cer} but with a larger coverage gap. Except for \gls{seqcp}, the other baselines obtain good coverage in most settings at the expense of drastically wider \glspl{pi}. Regarding computational scalability, note that \gls{method} shares most of the learnable parameters among time series and that its training can be efficiently parallelized on a GPU. Furthermore, besides learnable node embeddings $\mV$, \gls{method} relies only on a short window of the most recent observations as an input at each time step. Conversely, \gls{scpi} requires training a different model for each time series, and \gls{hopcpt} requires computing attention scores w.r.t.\ the entire calibration set at each time step.  

\subsection{Controlled Environment}\label{sec:gpvar-exp}

\input{tables/table-gpvar}

We evaluated the behavior of \gls{method} in a controlled environment by simulating a diffusion process on a graph. In particular, the experiment relied on the \gls{gpvar} benchmark introduced by \citet{zambon2022aztest}, with a setup analogous to \cite{cini2023taming}.
Data were generated recursively from the auto-regressive polynomial graph filter:
\begin{align}\label{eq:gpvar}
    \mH_t &= \sum_{l=1}^L\sum_{q=1}^{Q} \Theta_{q,l}\mA^{l-1}\mX_{t-q},\notag\\
    \mX_{t+1} &= a \odot \text{tanh}\left(\mH_t\right) + b\odot\text{tanh}\left(\mX_{t-1}\right) + \eta_t,
\end{align}
where parameters $\boldsymbol{\Theta}\in\sR^{Q \times L}$, $a\in\sR$, $b\in\sR$ are kept fixed across nodes and $\eta_t \sim \mathcal{N}(\boldsymbol{0}, \sigma^2\sI)$ with $\sigma = 0.4$. We ran the simulation on a graph with $60$ nodes and with a topology analogous to previous works~\cite{cini2023taming}; more details are provided in \autoref{a:datasets}. 
We use both a \gls{rnn} and \gls{stgnn} as base point predictors. As shown by \citet{zambon2022aztest}, \glspl{stgnn} can obtain a forecasting accuracy near the theoretical optimum in this dataset, which results in uncorrelated residuals. As such, we would expect standard \gls{scp} to be sufficient when using an \gls{stgnn} as base model. The objective of this experiment is to show that \gls{method} is effectively able to capture and leverage existing spatiotemporal dependencies.

\paragraph{Results} We compare \gls{method} against standard \gls{scp} and the \gls{cornn} variant. Moreover, we also compare performance against \gls{method} with access to the true graph used to generate the data. Results are shown in \autoref{t:gpvar}. When using an \gls{rnn} as a point predictor~(\textbf{\gls{gpvar}-\gls{rnn}}) \gls{method} significantly outperforms both standard \gls{scp} and \gls{cornn}. 
Furthermore, \gls{method} achieves results that closely match those obtained with direct access to the ground truth graph, which shows the effectiveness of the proposed architecture in capturing latent relational dependencies. Note that, given the injected Gaussian noise with $\sigma=0.4$, the theoretical optimum \gls{pi} width to obtain~(asymptotically) $90\%$ marginal coverage is $1.315$. \Gls{method} achieves essentially perfect coverage with PI width close to the theoretical optimum, while \gls{cornn} requires a substantially higher \gls{pi} width to obtain similar coverage.  Finally, results obtained by using an \gls{stgnn} as baseline~(\textbf{\gls{gpvar}-\gls{stgnn}}) show~(as expected) that the standard \gls{scp} is sufficient when the point-predictor captures all the relevant dependencies. We provide a visualization of the learned graph in \autoref{a:gpvar}. 

\subsection{Adaptation}\label{sec:adpatation-exp}

\input{tables/table-adaptation}

In this experiment, we evaluate how effectively the adaptation technique proposed in \autoref{sec:adaptive-inference} provides accurate \glspl{pi} over time. We focused on the \gls{cer} dataset where the calibration set does not cover a full year, which likely introduces a shift at test time given the seasonality of energy consumption. 
We used \gls{method} with a fixed hyperparameter configuration across scenarios and trained on the calibration set. Embeddings are updated every $M$ time steps by running the training procedure on the latest observations and keeping all the parameters frozen except for the embeddings. More details on the hyperparameters are provided in \autoref{a:exp}. 
In practice, this is done by splitting the test set into $K=6$ folds and then iteratively fine-tuning the model on each fold. This procedure simulates a real-world scenario where new data become available over time and are used for fine-tuning.  Results, reported in \autoref{t:adaptation}, show that this adaptation scheme improves performance by reducing the coverage gap and providing more accurate \glspl{pi}. 

\section{Conclusion}\label{sec:conclusion}

In this paper, we introduced \acrfull{method}, a novel \gls{cp} method for correlated time series. \gls{method} exploits graph-based neural operators to implement an uncertainty quantification architecture that can operate on top of any pre-trained point predictor. Furthermore, our approach does not require the relational structure to be known in advance.  Results show that the proposed method compares favorably against the state-of-the-art in several relevant scenarios. 

\paragraph{Limitations and future work} We believe that \gls{method} constitutes an important step toward effective spatiotemporal \gls{cp} methods. There are several directions for future work to explore, such as applying the framework to heterogeneous time series. Future work should focus on mitigating challenges arising from learning in non-stationary environments, e.g., by developing methods and coverage guarantees under specific assumptions about the non-stationarity of the data-generating process. Applications of \gls{method} to very large time series collections and combinations thereof with scalable graph-based architectures~\cite{chiang2019cluster, cini2023sparse} are an additional direction. Finally, while \gls{method} accounts for dependencies w.r.t.\ past observations at correlated time series, it provides separate \glspl{pi} for each time series. It would then be interesting to extend the framework with components that model the joint probability distribution of the time series.

%% file: imgs/figure_1.tex
\begin{figure*}
    \centering
    \includegraphics[width=\textwidth]{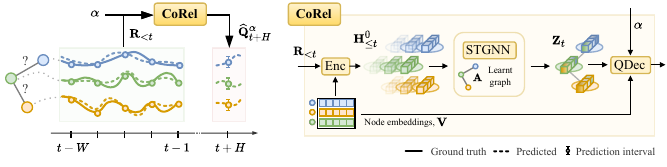}
    \caption{Overview of \gls{method}. Past residuals are used as input to a hybrid global-local graph-based quantile network.}
    \label{fig:corel}
\end{figure*}

%% file: tables/table-benchmarks.tex
\begin{table*}[t]
\centering

\setlength{\tabcolsep}{4pt}
\setlength{\aboverulesep}{0pt}
\setlength{\belowrulesep}{0pt}
\renewcommand{\arraystretch}{1.1}

\begin{tabular}{@{}l|l|l|ccccc|cc@{}}
                                \multicolumn{2}{c}{}    & Metric       & \gls{scp} & \gls{nexcp} & \gls{seqcp} & \gls{scpi} & \gls{hopcpt} & \gls{cornn} & \gls{method} \\
\midrule[1.5pt]
\multirow{9}{*}{\rotatebox{90}{\gls{la}}}
 & \multirow{3}{*}{\rotatebox{90}{RNN}} & $\Delta$Cov  & \textcolor{tabgreen}{-1.28} & \textcolor{tabgreen}{-1.00} & \textcolor{tabred}{-6.93} & \textcolor{tabgreen}{-1.24{\tiny$\pm$0.01}} & \textcolor{tabgreen}{-1.46{\tiny$\pm$0.11}} & \textcolor{tabgreen}{-0.49{\tiny$\pm$0.53}} & \textcolor{tabgreen}{-1.29{\tiny$\pm$0.21}} \\ &                             & PI-Width     & 20.69 & 25.23 & 19.52 & 19.84{\tiny$\pm$0.01} & 16.62{\tiny$\pm$0.10} & 19.48{\tiny$\pm$0.43} & 14.38{\tiny$\pm$0.22} \\ &                             & Winkler      & 40.80 & 41.33 & 50.12 & 37.95{\tiny$\pm$0.01} & 25.63{\tiny$\pm$0.19} & 30.24{\tiny$\pm$0.18} & \textbf{23.78{\tiny$\pm$0.20}} \\
\cmidrule[0.2pt]{3-10}
 & \multirow{3}{*}{\rotatebox{90}{\textsc{Transf}}} & $\Delta$Cov  & \textcolor{tabgreen}{-1.18} & \textcolor{tabgreen}{-0.97} & \textcolor{tabred}{-6.98} & \textcolor{tabgreen}{-1.13{\tiny$\pm$0.00}} & \textcolor{tabgreen}{-1.22{\tiny$\pm$0.56}} & \textcolor{tabgreen}{-0.55{\tiny$\pm$0.61}} & \textcolor{tabgreen}{-1.02{\tiny$\pm$0.63}} \\ &                             & PI-Width     & 20.81 & 25.30 & 19.44 & 19.86{\tiny$\pm$0.01} & 16.66{\tiny$\pm$0.13} & 19.38{\tiny$\pm$0.64} & 14.17{\tiny$\pm$0.43} \\ &                             & Winkler      & 40.55 & 41.64 & 49.90 & 37.74{\tiny$\pm$0.01} & 25.37{\tiny$\pm$0.33} & 30.33{\tiny$\pm$0.18} & \textbf{23.65{\tiny$\pm$0.19}} \\
\cmidrule[0.2pt]{3-10}
 & \multirow{3}{*}{\rotatebox{90}{STGNN}} & $\Delta$Cov  & \textcolor{tabgreen}{-0.99} & \textcolor{tabgreen}{-0.62} & \textcolor{tabred}{-13.60} & \textcolor{tabgreen}{-0.87{\tiny$\pm$0.00}} & \textcolor{tabgreen}{-0.13{\tiny$\pm$0.27}} & \textcolor{tabgreen}{-0.34{\tiny$\pm$0.49}} & \textcolor{tabgreen}{-0.92{\tiny$\pm$0.31}} \\ &                             & PI-Width     & 17.30 & 22.20 & 12.87 & 16.38{\tiny$\pm$0.01} & 15.75{\tiny$\pm$0.19} & 16.20{\tiny$\pm$0.24} & 14.65{\tiny$\pm$0.24} \\ &                             & Winkler      & 34.94 & 34.49 & 40.36 & 33.66{\tiny$\pm$0.01} & \textbf{22.80{\tiny$\pm$0.28}} & 28.74{\tiny$\pm$0.12} & 24.70{\tiny$\pm$0.16} \\
\midrule[1.5pt]
\multirow{9}{*}{\rotatebox{90}{\gls{cer}}}
 & \multirow{3}{*}{\rotatebox{90}{RNN}} & $\Delta$Cov  & \textcolor{taborange}{-3.46} & \textcolor{tabgreen}{0.11} & \textcolor{taborange}{-3.57} & \textcolor{taborange}{-3.45{\tiny$\pm$0.00}} & \textcolor{tabred}{-4.37{\tiny$\pm$0.32}} & \textcolor{tabolive}{-2.24{\tiny$\pm$0.34}} & \textcolor{taborange}{-3.60{\tiny$\pm$0.35}} \\ &                             & PI-Width     & 2.60 & 3.26 & 2.75 & 2.39{\tiny$\pm$0.00} & 1.97{\tiny$\pm$0.03} & 1.96{\tiny$\pm$0.02} & 1.83{\tiny$\pm$0.03} \\ &                             & Winkler      & 5.69 & 5.48 & 5.79 & 5.29{\tiny$\pm$0.00} & 3.87{\tiny$\pm$0.07} & 3.84{\tiny$\pm$0.01} & \textbf{3.71{\tiny$\pm$0.04}} \\
\cmidrule[0.2pt]{3-10}
 & \multirow{3}{*}{\rotatebox{90}{\textsc{Transf}}} & $\Delta$Cov  & \textcolor{taborange}{-3.35} & \textcolor{tabgreen}{0.10} & \textcolor{taborange}{-3.53} & \textcolor{taborange}{-3.26{\tiny$\pm$0.01}} & \textcolor{taborange}{-3.95{\tiny$\pm$0.60}} & \textcolor{tabolive}{-2.04{\tiny$\pm$0.32}} & \textcolor{taborange}{-3.97{\tiny$\pm$0.24}} \\ &                             & PI-Width     & 2.52 & 3.16 & 2.67 & 2.33{\tiny$\pm$0.00} & 2.01{\tiny$\pm$0.09} & 1.94{\tiny$\pm$0.03} & 1.80{\tiny$\pm$0.02} \\ &                             & Winkler      & 5.60 & 5.36 & 5.69 & 5.20{\tiny$\pm$0.00} & 3.88{\tiny$\pm$0.12} & 3.82{\tiny$\pm$0.02} & \textbf{3.67{\tiny$\pm$0.02}} \\
\cmidrule[0.2pt]{3-10}
 & \multirow{3}{*}{\rotatebox{90}{STGNN}} & $\Delta$Cov  & \textcolor{tabred}{-4.30} & \textcolor{tabgreen}{0.08} & \textcolor{taborange}{-3.83} & \textcolor{tabred}{-4.17{\tiny$\pm$0.01}} & \textcolor{tabred}{-5.06{\tiny$\pm$0.15}} & \textcolor{tabolive}{-2.13{\tiny$\pm$0.83}} & \textcolor{tabred}{-4.99{\tiny$\pm$0.61}} \\ &                             & PI-Width     & 2.28 & 3.00 & 2.42 & 2.09{\tiny$\pm$0.00} & 1.79{\tiny$\pm$0.01} & 1.85{\tiny$\pm$0.04} & 1.77{\tiny$\pm$0.05} \\ &                             & Winkler      & 5.11 & 4.87 & 4.99 & 4.76{\tiny$\pm$0.00} & \textbf{3.49{\tiny$\pm$0.02}} & 3.72{\tiny$\pm$0.02} & 3.76{\tiny$\pm$0.04} \\
\midrule[1.5pt]
\multirow{9}{*}{\rotatebox{90}{\gls{air}}}
 & \multirow{3}{*}{\rotatebox{90}{RNN}} & $\Delta$Cov  & \textcolor{tabgreen}{5.09} & \textcolor{tabgreen}{-0.63} & \textcolor{taborange}{-3.21} & \textcolor{tabgreen}{1.79{\tiny$\pm$0.01}} & \textcolor{tabgreen}{0.01{\tiny$\pm$2.62}} & \textcolor{tabolive}{-2.06{\tiny$\pm$1.52}} & \textcolor{tabolive}{-2.78{\tiny$\pm$0.60}} \\ &                             & PI-Width     & 118.06 & 82.04 & 74.73 & 103.88{\tiny$\pm$0.03} & 90.35{\tiny$\pm$13.04} & 71.61{\tiny$\pm$1.82} & 68.13{\tiny$\pm$1.30} \\ &                             & Winkler      & 148.61 & 131.18 & 135.59 & 143.10{\tiny$\pm$0.01} & 133.24{\tiny$\pm$8.31} & 113.11{\tiny$\pm$0.77} & \textbf{107.67{\tiny$\pm$0.94}} \\
\cmidrule[0.2pt]{3-10}
 & \multirow{3}{*}{\rotatebox{90}{\textsc{Transf}}} & $\Delta$Cov  & \textcolor{tabgreen}{5.07} & \textcolor{tabgreen}{-0.66} & \textcolor{taborange}{-3.17} & \textcolor{tabgreen}{1.60{\tiny$\pm$0.01}} & \textcolor{tabolive}{-2.34{\tiny$\pm$1.19}} & \textcolor{tabgreen}{-1.93{\tiny$\pm$0.44}} & \textcolor{tabgreen}{-1.81{\tiny$\pm$1.67}} \\ &                             & PI-Width     & 118.50 & 81.15 & 75.08 & 104.11{\tiny$\pm$0.01} & 80.37{\tiny$\pm$2.09} & 73.25{\tiny$\pm$0.95} & 72.25{\tiny$\pm$3.16} \\ &                             & Winkler      & 150.33 & 132.08 & 137.62 & 145.14{\tiny$\pm$0.02} & 129.85{\tiny$\pm$3.14} & 112.71{\tiny$\pm$0.44} & \textbf{108.71{\tiny$\pm$1.38}} \\
\cmidrule[0.2pt]{3-10}
 & \multirow{3}{*}{\rotatebox{90}{STGNN}} & $\Delta$Cov  & \textcolor{tabgreen}{4.48} & \textcolor{tabgreen}{-0.32} & \textcolor{tabolive}{-2.94} & \textcolor{tabgreen}{2.64{\tiny$\pm$0.01}} & \textcolor{tabgreen}{-0.78{\tiny$\pm$1.34}} & \textcolor{tabgreen}{-1.52{\tiny$\pm$0.51}} & \textcolor{tabgreen}{-2.00{\tiny$\pm$1.54}} \\ &                             & PI-Width     & 111.68 & 80.01 & 71.92 & 99.90{\tiny$\pm$0.03} & 79.39{\tiny$\pm$3.38} & 70.17{\tiny$\pm$0.80} & 68.24{\tiny$\pm$2.65} \\ &                             & Winkler      & 143.39 & 127.98 & 130.86 & 137.09{\tiny$\pm$0.03} & 121.26{\tiny$\pm$2.48} & 109.65{\tiny$\pm$0.30} & \textbf{108.45{\tiny$\pm$1.52}} \\
\bottomrule[1.5pt]
\end{tabular}

\caption{Performance comparison for $\alpha=0.1$. $\Delta$Cov values are color-coded: \textcolor{tabgreen}{green} (0-2\%), \textcolor{tabolive}{yellow} (2-3\%), \textcolor{taborange}{orange} (3-4\%), \textcolor{tabred}{red} ($>$4\%). The lowest Winkler score for each scenario is shown in bold.}
\label{tab:exp1}
\end{table*}

%% file: tables/table-gpvar.tex
\begin{table}[t]
\caption{Performance on GPVAR.}
\label{t:gpvar}
\small
\setlength{\tabcolsep}{1.2pt}
\setlength{\aboverulesep}{0.2pt}
\setlength{\belowrulesep}{0.2pt}
\renewcommand{\arraystretch}{1.2}
\begin{center}
\begin{tabular}{c|c|c c c | c c c}
 \multicolumn{2}{c|}{} & \multicolumn{3}{c|}{\gls{gpvar}-\gls{rnn}} & \multicolumn{3}{c}{\gls{gpvar}-\gls{stgnn}} \\
  \cmidrule{3-8}
 \multicolumn{2}{c|}{Models} & \multicolumn{1}{c}{$\Delta$Cov} &\multicolumn{1}{c}{PI-Width} &\multicolumn{1}{c|}{Winkler} & \multicolumn{1}{c}{$\Delta$Cov} &\multicolumn{1}{c}{PI-Width} &\multicolumn{1}{c}{Winkler} \\
\midrule
\multicolumn{2}{c|}{\gls{scp}} &-0.02 & 1.67 & 2.14 & -0.02 & 1.32 & 1.66 \\
\midrule
\multicolumn{2}{c|}{\gls{cornn}} &-0.1{\tiny $\pm$0.2} & 1.63{\tiny $\pm$0.01} & 2.04{\tiny$\pm$0.00} & 0.0{\tiny $\pm$0.0} & 1.32{\tiny$\pm$0.00} & 1.66{\tiny$\pm$0.00} \\
\midrule
\multicolumn{2}{c|}{\gls{method}} &0.0{\tiny $\pm$0.1} & 1.33{\tiny$\pm$0.00} & 1.67{\tiny$\pm$0.00}&0.1{\tiny $\pm$0.1} & 1.32{\tiny $\pm$0.01} & 1.66{\tiny$\pm$0.00}\\
\multicolumn{2}{c|}{w/ true $\mA$} &-0.1{\tiny $\pm$0.1} & 1.32{\tiny$\pm$0.00} & 1.66{\tiny$\pm$0.00}&0.0{\tiny $\pm$0.2} & 1.32{\tiny $\pm$0.01} & 1.66{\tiny$\pm$0.00} \\
\bottomrule
\end{tabular}
\end{center}
\end{table}

%% file: tables/table-adaptation.tex
\begin{table*}[!ht]
\centering
\caption{Adaptation results for \textbf{\gls{cer}} for a reference \gls{method} model. ($\alpha=0.1$).}
\label{t:adaptation}
\setlength{\tabcolsep}{4pt}
\setlength{\aboverulesep}{1pt}
\setlength{\belowrulesep}{1pt}
\begin{tabular}{l ccc ccc ccc}
\toprule
& \multicolumn{3}{c}{\textbf{RNN}} 
& \multicolumn{3}{c}{\textbf{Transformer}} 
& \multicolumn{3}{c}{\textbf{STGNN}} \\
\cmidrule(lr){2-4} \cmidrule(lr){5-7} \cmidrule(lr){8-10}
 
& \(\Delta\)Cov & PI-Width & Winkler 
& \(\Delta\)Cov & PI-Width & Winkler 
& \(\Delta\)Cov & PI-Width & Winkler \\
\midrule
W/o adapt.\
& \textcolor{taborange}{-3.44 {\tiny $\pm$ 0.40}} & 1.84 {\tiny $\pm$ 0.02} & 3.71 {\tiny $\pm$ 0.03}
& \textcolor{taborange}{-3.08 {\tiny $\pm$ 0.15}} & 1.84 {\tiny $\pm$ 0.02} & 3.69 {\tiny $\pm$ 0.04}
& \textcolor{tabred}{-4.07 {\tiny $\pm$ 0.47}} & 1.80 {\tiny $\pm$ 0.03} & 3.78 {\tiny $\pm$ 0.03} \\
W/ adapt.\
& \textcolor{tabolive}{-2.70 {\tiny $\pm$ 0.27}} & 1.85 {\tiny $\pm$ 0.01} & \textbf{3.49 {\tiny $\pm$ 0.02}}
& \textcolor{tabolive}{-2.42 {\tiny $\pm$ 0.20}} & 1.85 {\tiny $\pm$ 0.01} & \textbf{3.49 {\tiny $\pm$ 0.04}}
& \textcolor{taborange}{-3.06 {\tiny $\pm$ 0.25}} & 1.81 {\tiny $\pm$ 0.02} & \textbf{3.58 {\tiny $\pm$ 0.02}}\\
\bottomrule
\end{tabular}
\end{table*}

%% file: impact_statement.tex
\section*{Impact Statement}

This paper presents work whose goal is to advance the field of machine learning and time series forecasting. There are many potential societal consequences of our work, none of which we feel must be specifically highlighted here.

%% file: acks.tex
\section*{Acknowledgments}

This work was supported by the Swiss National Science Foundation grant no.~225351~(\emph{Relational Deep Learning for Reliable Time Series Forecasting at Scale}) and no.~204061~(\emph{HORD GNN: Higher-Order Relations and Dynamics in Graph Neural Networks}), the Hasler Stiftung project \emph{Calibrated Uncertainty Estimation for Spatio-Temporal Data}, the UKRI CDT in AI for Healthcare (grant no. P/S023283/1), and the Norwegian Research Council grant no.~345017~(\emph{RELAY: Relational Deep Learning for Energy Analytics}). FMB wishes to thank Nvidia Corporation for donating some of the GPUs used in this project. AC conducted part of this work while at the University of Oxford.

%% file: appendix.tex
\section*{Appendix}

\section{Proof of \autoref{p:coverage-bound}}\label{a:theory}

As stated in \autoref{p:coverage-bound}, let $P^{c}_{t+H}$ and $P^c_{\psi}$ indicate the probability distributions associated with the true data distribution at $t+H$ and with the learned quantile function $\gQ_{\psi}$, respectively. It follows from the definition of total variation distance $\,{TV}(P,Q) := \sup_{B}|P(B)-Q(B)|,$ that for any event $B$
\begin{align}
    &|P^{c}_{t+H}(B)-P^c_{\psi}(B)|\,\le\,{{TV}}(P^c_{\psi},P^{c}_{t+H}) \quad \Longrightarrow \\
    &P^{c}_{t+H}(B)\,\ge\,P^c_{\psi}(B)\,-\,{{TV}}(P^c_{\psi},P^c_{t+H}).\label{eq:tv-bound}
\end{align}
For $ \hat C^{\alpha}_{i,t}$ in both \autoref{eq:cp-interval} and \autoref{eq:cp-interval-optimized} we have by construction that
\begin{equation}\label{eq:proof-quantiles}
    P^c_{\psi}\left(\vx^i_{t+H} \in \widehat C^{\alpha}_{i,t}\right) = 1 - \alpha.
\end{equation}
Then, putting together \autoref{eq:tv-bound} and \autoref{eq:proof-quantiles} we obtain
\begin{align*}
    P^c_{t+H}\left(\vx^i_{t+H} \in \widehat C^{\alpha}_{i,t} (\widehat \mX_{t+h})\right) \geq 1 - \alpha - {TV}\left(P^c_{\psi}, P^c_{t+H}\right).
\end{align*}
Hence, the proof is complete.

\section{Hardware and software platforms}\label{a:exp}

Benchmarks have been developed with Python~\cite{rossum2009python} and the following open-source libraries:
\begin{itemize}
    \item Numpy~\cite{harris2020array};
    \item PyTorch~\cite{paske2019pytorch};
    \item PyTorch Lightning~\cite{Falcon_PyTorch_Lightning_2019};
    \item PyTorch Geometric~\cite{fey2019fast};
    \item Torch Spatiotemporal~\cite{Cini_Torch_Spatiotemporal_2022}.
\end{itemize}

Experiments were conducted on a server equipped with AMD EPYC 7513 CPUs and NVIDIA RTX A5000 GPUs. The code for reproducing the computational experiments is available at \repourl. For the \gls{hopcpt}, \gls{scpi}, \gls{nexcp}, and \gls{seqcp} baselines we use the open-source implementation made available by \citet{auer2023conformal}\footnote{\url{https://github.com/ml-jku/HopCPT}}. In our setup for the experiment in \autoref{sec:benchmarks}, considering the \gls{la} dataset and the baselines that require fitting a model, training and testing require $\approx 3$ days for \gls{scpi}, $\approx 11$ hours for \gls{hopcpt}, and $\approx 5$ minutes for \gls{method}.

\section{Datasets}\label{a:datasets}

We considered several datasets from different application domains, real-world scenarios and a simulated controlled environment. For the real world datasets we followed the pre-processing steps adopted by \citet{cini2023taming}.

\paragraph{Air Quality Monitoring} The \textbf{\gls{air}} \cite{zheng2015forecasting} dataset consists of 8,760 hourly measurements of pollutant PM2.5 from 437 monitoring stations in China. We use a window of $W=24$ time steps and predict the $3$-time-step-ahead observations. For the \gls{stgnn} base model, we build an adjacency matrix using a thresholded Gaussian kernel computed from pairwise geographic distances~\cite{shuman2013emerging}.

\paragraph{Traffic Forecasting} We considered the \textbf{\gls{la}} \cite{li2018diffusion} traffic forecasting dataset, consisting of 34,272 timesteps of measurements from 207 loop detectors sampled at 5-minute intervals in the Los Angeles County Highways. We use a window of $12$ time steps and predict the $12$-time-step-ahead observations. For the \gls{stgnn} base model, we followed previous works~\cite{wu2019graph} and built an adjacency matrix using a thresholded Gaussian kernel applied to geographic distances.

\paragraph{Electric Load Forecasting} We selected the \textbf{\gls{cer}} dataset \cite{cer2016cer, cini2022filling}, comprising 25,728 timesteps of energy consumption readings aggregated at 30-minute intervals from 485 smart meters monitoring small and medium-sized enterprises. We use a window of $48$ time steps and predict the $5$-time-step-ahead observations. For the \gls{stgnn} base model, we built the adjacency matrix by extracting a $10$-nearest neighbor graph from week-wise correntropy similarities between time series, following previous work~\cite{cini2022filling}. 

\paragraph{GPVAR} For the \gls{gpvar} dataset, we generate synthetic data with 40,000 timesteps over an undirected network of 60 nodes connected in a community graph structure by following the system model in \autoref{eq:gpvar}~\cite{zambon2022aztest}. The parameters of the spatiotemporal process are set as
\begin{equation}
    \Theta = \begin{bmatrix} 2.5 & -2.0 & -0.5 \\ 1.0 & 3.0 & 0.0 \end{bmatrix}, \quad a=b=0.5, \quad \sigma = 0.4.
\end{equation}
We used an input window of $5$ time steps to predict the next observation. For the \gls{stgnn} base model, we used the same community graph structure as the adjacency matrix. 

\paragraph{Base models} We trained three base models~(point predictors) for each dataset: a \gls{rnn} with \gls{gru} cells (1 layer with hidden size 32), a decoder-only Transformer (hidden size 32, feed-forward size 64, 2 attention heads, 3 layers, dropout 0.1), and a \gls{stgnn} following the template in \autoref{sec:stgnn} (hidden size 32, node embedding size 16, 1 layer \gls{gru}, 2 message-passing layers). All models were trained by minimizing the MAE loss using the Adam optimizer for 200 epochs with batch size 32, using 40\% of the data for training, 40\% for calibration, and 20\% for testing. We also use the first $25\%$ of the calibration data as a validation set for early stopping. Input features were scaled using standard scaling across time series. 

\section{Additional details on \gls{method} implementation}\label{a:corel}

\paragraph{Architecture} As discussed in \autoref{sec:method} and \autoref{sec:experiments}, we implemented \gls{method} as 
\gls{tts} model with a single-layer \gls{gru} followed by $2$ message passing layers analogous to those in~\cite{satorras2022multivariate}. As a readout, we used an \gls{mlp} mapping the learned representations to predictions of $39$ equally spaced quantiles. For the \gls{cornn} baseline, we used the same architecture but removed the message passing layers and node embeddings $\mV$ and allowed the \gls{gru} to have more than one layer.

\paragraph{Latent graph learning module} The graph learning module was implemented as described in \autoref{sec:structure-learning} by parametrizing $\mPhi$ with a matrix of $N\times N$ learnable parameters. To allow for sampling less neighbors than the specified neighborhood size $K$, we followed previous works~\cite{cini2023sparse} and modified the sampling procedure by introducing a set of dummy nodes then discarded from the sampled graph before message passing. We use dummy nodes in the \gls{gpvar} experiment only. To allow for sparse message-passing operations we use a straight-through gradient estimator~\cite{bengio2013estimating} and backpropagate gradients only through the sampled edges for each node, plus $10\%$ of the remaining ones. In the \gls{gpvar} experiment, we simply propagate gradients w.r.t.\ the entire adjacency matrix. 

\paragraph{Covariates} Besides residuals, we use as additional covariates datetime encodings whenever available, plus the value of the target time series w.r.t.\ the time steps in the input window.

\section{Evaluation metrics}
For a prediction interval $\widehat{C}_i^\alpha = [\hat{x}_i + \hat{q}_i^{\alpha/2}, \hat{x}_i + \hat{q}_i^{1-\alpha/2}]$ and true value $x_i$, evaluation is conducted for a desired confidence level $\alpha$ using three key metrics. 

\paragraph{Coverage gap} $\Delta\text{Cov}_i$ measures the difference between the achieved coverage and the target coverage $1-\alpha$, and is given by
\begin{equation}
    \text{$\Delta$Cov}_i = 100\left(\mathbbm{1}(x_i \in \widehat{C}_i^\alpha) - (1-\alpha)\right),
\end{equation}
where $\mathbbm{1}$ denotes the indicator function.

\paragraph{Prediction interval width} Quantifies the width of the prediction intervals, and is given by
\begin{equation}
    \text{PI-Width}_i = \hat{q}_i^{1-\alpha/2} - \hat{q}_i^{\alpha/2}.
\end{equation}

\paragraph{Winkler Score} Combines interval width with a penalty for predictions that fail to capture the true value, with misses penalized by a factor of $\frac{2}{\alpha}$. The Winkler score is given by
\begin{equation}
    W_i = \begin{cases}
        (\hat{q}^{1-\alpha/2}_i - \hat{q}^{\alpha/2}_i) + \frac{2}{\alpha}(\hat{q}^{\alpha/2}_i - x_i) & \text{if } x_i < \hat{q}^{\alpha/2}_i, \\
        (\hat{q}^{1-\alpha/2}_i - \hat{q}^{\alpha/2}_i) & \text{if } \hat{q}^{\alpha/2}_i \leq x_i \leq \hat{q}^{1-\alpha/2}_i, \\
        (\hat{q}^{1-\alpha/2}_i - \hat{q}^{\alpha/2}_i) + \frac{2}{\alpha}(x_i - \hat{q}^{1-\alpha/2}_i) & \text{if } x_i > \hat{q}^{1-\alpha/2}_i.
    \end{cases}
\end{equation}
All the metrics are then averaged over all nodes and time steps within the specified set.

\section{Hyperparameters and experimental setup}

Hyperparameters were tuned separately for each combination of base predictor and dataset on a validation set. 

\paragraph{\gls{method}} For \gls{method} we tuned the number of neurons in the \gls{stgnn} with a small grid search on $10\%$ of the calibration data. We used the same model selection procedure for \gls{cornn} but also tuned the number of \gls{gru} layers. For the experiments on real-world data, the model was trained for a maximum of $100$ epochs on the calibration set. Each epoch consisted of a maximum of $50$ mini-batches of size $64$. We used the Adam optimizer~\cite{kingma2014adam} with an initial learning rate of $0.003$ and reduced by $75\%$ every $20$ epochs. We used a fixed number of $K=20$ neighbors for the graph learning module. For the \gls{gpvar} experiment in \autoref{sec:gpvar-exp}, we fixed the number of neurons to $16$ for each layer and used an embedding size of $8$. For the adaptation experiment in \autoref{sec:adpatation-exp}, we use a reference configuration with $64$ neurons in each encoder/decoder layers and an embedding size of $32$. We train the entire model on the full calibration set and use adaptation at test time. In particular, at test time, we fine-tune the node embeddings every $M$ time steps, where $M$ corresponds to $\frac{1}{6}$ of the test set length. For fine-tuning, we fit the embeddings by running $25$ epochs of maximum $10$ batches each by using samples from the last $M$ steps with a fixed learning rate of $0.001$. 

\paragraph{\gls{hopcpt}}  For \gls{hopcpt} we followed the model selection procudere described in~\cite{auer2023conformal}. 
The model was trained for $3000$ epochs using a batch size of $4$ time series. We adopted the paper's AdamW optimizer configuration with standard parameters ($\beta_1 = 0.9$, $\beta_2 = 0.999$, $\epsilon = 0.01$) and tuned the model by running the same hyperparameter configurations searched by \cite{auer2023conformal}. All the remaining hyperparameters were set accordingly to the original paper. 

\paragraph{\gls{seqcp}} \gls{seqcp}, similarly to \cite{xu2023conformal}, employs a sliding window approach to conformal prediction, assigning equal weights to observations within the most recent $K$ time steps and zero weights to older observations. The window size $K$ was treated as a hyperparameter and tuned over the values $\{$200, 150, 125, 100, 75, 50, 25, 10$\}$.

\paragraph{\gls{nexcp}} \gls{nexcp} implements conformal prediction using exponentially decaying weights controlled by a parameter $\rho$. Rather than using a fixed window of historical observations, it assigns weights that decay exponentially with time, giving more recent observations higher importance. The decay parameter $\rho$ was tuned over the values $\{$0.999, 0.995, 0.993, 0.99, 0.98, 0.95, 0.9$\}$.

\paragraph{\gls{scpi}} As already mentioned, we used the \gls{scpi}~\cite{xu2023sequential} implementation provided by \citet{auer2023conformal} and followed an analogous protocol for training the mode. In particular, \gls{scpi} was run using a fixed window length of $100$ time steps for all experiments, corresponding to the longest setting in the original paper. The computational demands of SPCI were substantial, as separate quantile random forests had to be trained for each combination of node and target coverage level, making extensive hyperparameter tuning impractical. To further manage the computational costs, we trained each SPCI model only once on the calibration data, rather than implementing the time-adaptive approach where models are re-trained as new observations become available. 

\section{Additional results}

\begin{figure}[t]
    \centering
    \includegraphics[width=.85\textwidth]{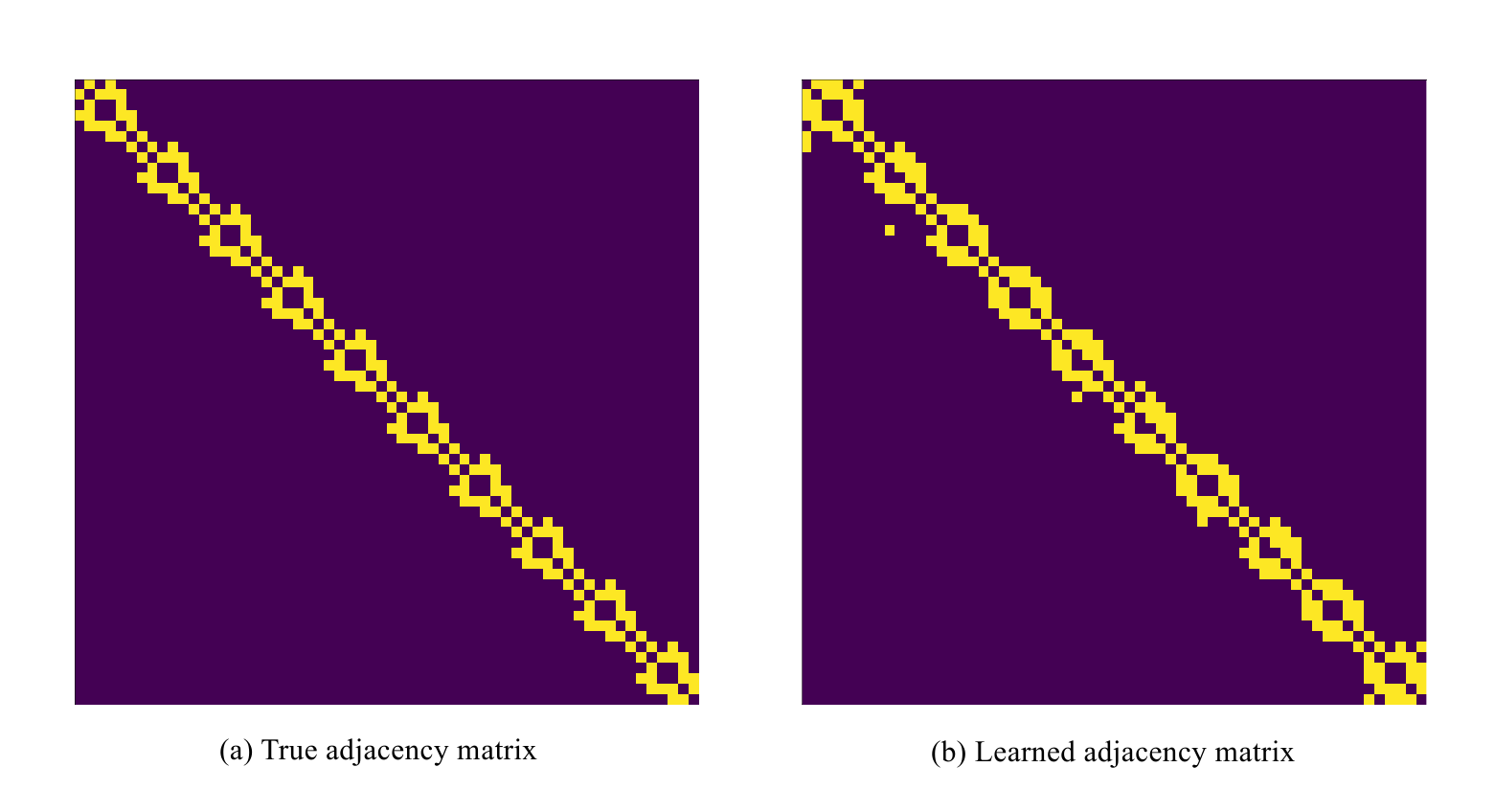}
    \caption{Comparison between \textbf{(a)} the true graph and \textbf{(b)} the graph structure learned by \gls{method} in \gls{gpvar}.}
    \label{fig:gpvar-graph}
\end{figure}

\subsection{Qualitative analysis of the learned graph on \gls{gpvar}}\label{a:gpvar}

\autoref{fig:gpvar-graph} provides a comparison of the graph learned on \gls{gpvar} by \gls{method} in \autoref{sec:gpvar-exp}. In particular, the rightmost figure is obtained by taking the top-$K$ scores associated to each row\footnote{The figure does not show edge scores associated to dummy nodes; see \autoref{a:corel}}. The figure shows that the learned graph includes true edges plus additional links. Note that perfectly recovering the ground truth graph is not required to achieve optimal performance here.

\subsection{Optimization of the \gls{pi} width}\label{a:beta}

Here, we show how the width of the \gls{pi} can be reduced by searching for an appropriate quantile offset $\beta$.
Specifically, we perform the optimization described in \autoref{eq:cp-interval-optimized}, which identifies pairs of quantiles yielding the same coverage $1-\alpha$ while achieving the smallest interval width. 
We summarize our results in \autoref{t:beta_optim}. 
As we can see, the procedure consistently finds intervals with a smaller width at the price of slightly reducing the coverage, whether using \gls{method} or the \gls{cornn} variant.

\input{tables/table-beta-optim}

%% file: tables/table-beta-optim.tex
\begin{table*}[!ht]
\centering
\caption{Changes in $\Delta$ Cov and PI-Width when optimizing $\beta$ ($\alpha=0.1$).}
\label{t:beta_optim}
\setlength{\tabcolsep}{4pt}
\setlength{\aboverulesep}{1pt}
\setlength{\belowrulesep}{1pt}
\begin{tabular}{lll cc cc cc}
\toprule
& & & \multicolumn{2}{c}{\textbf{RNN}} & \multicolumn{2}{c}{\textbf{Transformer}} & \multicolumn{2}{c}{\textbf{STGNN}} \\
\cmidrule(lr){4-5} \cmidrule(lr){6-7} \cmidrule(lr){8-9}
& & & $\Delta$Cov & PI-Width & $\Delta$Cov & PI-Width & $\Delta$Cov & PI-Width \\
\cmidrule(lr){2-9}
\multirow{4}{*}{\rotatebox{90}{METR-LA}}& \multirow{2}{*}{\gls{cornn}} & & -0.70{ \tiny $\pm$ 0.20} & 19.69{ \tiny $\pm$ 0.27} & -0.52{ \tiny $\pm$ 0.56} & 19.07{ \tiny $\pm$ 0.42} & -0.67{ \tiny $\pm$ 0.47} & 16.02{ \tiny $\pm$ 0.13} \\
& & $\beta$ optim.\  & -0.80{ \tiny $\pm$ 0.28} & 17.71{ \tiny $\pm$ 0.28} & -0.96{ \tiny $\pm$ 0.35} & 17.16{ \tiny $\pm$ 0.46} & -0.88{ \tiny $\pm$ 0.44} & 14.98{ \tiny $\pm$ 0.17} \\
\cmidrule(lr){2-9}& \multirow{2}{*}{\gls{method}} & & -1.25{ \tiny $\pm$ 0.80} & 14.24{ \tiny $\pm$ 0.53} & -0.97{ \tiny $\pm$ 0.51} & 14.43{ \tiny $\pm$ 0.38} & -1.31{ \tiny $\pm$ 0.48} & 14.36{ \tiny $\pm$ 0.20} \\
& & $\beta$ optim.\  & -1.49{ \tiny $\pm$ 0.56} & 13.23{ \tiny $\pm$ 0.46} & -1.26{ \tiny $\pm$ 0.42} & 13.47{ \tiny $\pm$ 0.35} & -1.44{ \tiny $\pm$ 0.45} & 13.46{ \tiny $\pm$ 0.20} \\
\midrule
\multirow{4}{*}{\rotatebox{90}{CER-E}}& \multirow{2}{*}{\gls{cornn}} & & -2.09{ \tiny $\pm$ 0.56} & 1.93{ \tiny $\pm$ 0.01} & -1.70{ \tiny $\pm$ 0.37} & 1.99{ \tiny $\pm$ 0.02} & -2.45{ \tiny $\pm$ 0.41} & 1.85{ \tiny $\pm$ 0.03} \\
& & $\beta$ optim.\  & -2.63{ \tiny $\pm$ 0.65} & 1.82{ \tiny $\pm$ 0.01} & -2.28{ \tiny $\pm$ 0.31} & 1.88{ \tiny $\pm$ 0.02} & -3.10{ \tiny $\pm$ 0.52} & 1.78{ \tiny $\pm$ 0.03} \\
\cmidrule(lr){2-9}& \multirow{2}{*}{\gls{method}} & & -3.86{ \tiny $\pm$ 0.53} & 1.83{ \tiny $\pm$ 0.02} & -3.81{ \tiny $\pm$ 0.30} & 1.83{ \tiny $\pm$ 0.02} & -4.87{ \tiny $\pm$ 0.51} & 1.79{ \tiny $\pm$ 0.04} \\
& & $\beta$ optim.\  & -4.16{ \tiny $\pm$ 0.59} & 1.75{ \tiny $\pm$ 0.02} & -4.10{ \tiny $\pm$ 0.21} & 1.75{ \tiny $\pm$ 0.02} & -5.13{ \tiny $\pm$ 0.51} & 1.72{ \tiny $\pm$ 0.03} \\
\midrule
\multirow{4}{*}{\rotatebox{90}{AQI}}& \multirow{2}{*}{\gls{cornn}} & & -1.45{ \tiny $\pm$ 0.61} & 73.88{ \tiny $\pm$ 1.17} & -1.08{ \tiny $\pm$ 0.22} & 74.02{ \tiny $\pm$ 0.74} & -1.50{ \tiny $\pm$ 0.63} & 70.42{ \tiny $\pm$ 0.94} \\
& & $\beta$ optim.\  & -1.90{ \tiny $\pm$ 0.54} & 70.17{ \tiny $\pm$ 1.22} & -1.10{ \tiny $\pm$ 0.25} & 70.95{ \tiny $\pm$ 0.68} & -1.38{ \tiny $\pm$ 0.62} & 68.24{ \tiny $\pm$ 0.85} \\
\cmidrule(lr){2-9}& \multirow{2}{*}{\gls{method}} & & -2.40{ \tiny $\pm$ 1.97} & 70.44{ \tiny $\pm$ 2.69} & -3.54{ \tiny $\pm$ 1.80} & 68.31{ \tiny $\pm$ 4.49} & -2.61{ \tiny $\pm$ 1.92} & 67.76{ \tiny $\pm$ 3.12} \\
& & $\beta$ optim.\  & -2.83{ \tiny $\pm$ 1.86} & 67.35{ \tiny $\pm$ 2.31} & -4.19{ \tiny $\pm$ 1.96} & 65.20{ \tiny $\pm$ 4.20} & -3.22{ \tiny $\pm$ 2.45} & 65.32{ \tiny $\pm$ 3.43} \\
\bottomrule
\end{tabular}
\end{table*}